\documentclass[pdflatex,sn-basic, Numbered]{sn-jnl}
\usepackage{amssymb}
\usepackage{xcolor}
\usepackage{multirow} 
\usepackage{caption}
\usepackage{tabularray}
\usepackage{xurl}
\usepackage{amsmath}
\usepackage{array}
\usepackage{url}
\usepackage{hyperref}
\usepackage{float}
\usepackage{orcidlink}

\usepackage{graphicx}%
\usepackage{amsmath,amsfonts}%
\usepackage{amsthm}%
\usepackage{mathrsfs}%
\usepackage{textcomp}%
\usepackage{manyfoot}%
\usepackage{booktabs}%
\usepackage{algorithm}%
\usepackage{algorithmicx}%
\usepackage{algpseudocode}%
\usepackage{listings}%
\usepackage{natbib}

\hypersetup{
    colorlinks=true,
    citecolor=blue,
    linkcolor=red,
    urlcolor=magenta,
}

\let\cline\cmidrule

\begin{document}

\title[Article Title]{Impact of ML Optimization Tactics on Greener Pre-Trained ML Models}

\author[1]{\fnm{Alexandra} \sur{González Álvarez} \orcidlink{0009-0003-7634-0343}}\email{alexandra.gonzalez.alvarez@upc.edu}
\author[1]{\fnm{Joel} \sur{Castaño} \orcidlink{0009-0005-6385-9255}}\email{joel.castano@upc.edu}
\author[1]{\fnm{Xavier} \sur{Franch} \orcidlink{0000-0001-9733-8830}}\email{xavier.franch@upc.edu}
\author*[1]{\fnm{Silverio} \sur{Martínez-Fernández} \orcidlink{0000-0001-9928-133X}}\email{silverio.martinez@upc.edu}

\affil[1]{\orgname{Universitat Polit\`ecnica de Catalunya}, \city{Barcelona}, \country{Spain}}

\abstract{
\textbf{Background}: Given the fast-paced nature of today's technology, which has surpassed human performance in tasks like image classification, visual reasoning, and English understanding, assessing the impact of Machine Learning (ML) on energy consumption is crucial. Traditionally, ML projects have prioritized accuracy over energy, creating a gap in energy consumption during model inference. 

\textbf{Aims}: This study aims to (i) analyze image classification datasets and pre-trained models, (ii) improve inference efficiency by comparing optimized and non-optimized models, and (iii) assess the economic impact of the optimizations.

\textbf{Method}: We conduct a controlled experiment to evaluate the impact of various PyTorch optimization techniques (dynamic quantization, \texttt{torch.compile}, local pruning, and global pruning) to 42 Hugging Face models for image classification. The metrics examined include GPU utilization, power and energy consumption, accuracy, time, computational complexity, and economic costs. The models are repeatedly evaluated to quantify the effects of these software engineering tactics.

\textbf{Results}: Dynamic quantization demonstrates significant reductions in inference time and energy consumption, making it highly suitable for large-scale systems. Additionally, \texttt{torch.compile} balances accuracy and energy. In contrast, local pruning shows no positive impact on performance, and global pruning's longer optimization times significantly impact costs. 

\textbf{Conclusions}: This study highlights the role of software engineering tactics in achieving greener ML models, offering guidelines for practitioners to make informed decisions on optimization methods that align with sustainability goals. 
}


\maketitle

\section{Introduction}\label{sec:introduction}

Green software engineering has been highlighted by Gartner as one of the top five strategic technology trends in software engineering for 2024 and beyond \cite{gartner2024}. Indeed, ``the use of compute-heavy workloads increases an organization’s carbon footprint, and generative AI-enabled applications are especially energy-intensive, so implementing green software engineering will help organizations prioritize their sustainability objectives'' \cite{gartner2024}. These compute-heavy workloads have been recognized by researchers \cite{verdecchia2023systematic}, and governments \cite{whitehouseExecutiveOrder}. In response, the energy efficiency of Machine Learning (ML) frameworks has gained attention, with recent studies comparing popular frameworks like TensorFlow, PyTorch, and Scikit-learn in terms of energy consumption, memory usage, and execution time \cite{ajel2022energy}. For instance, Georgiou et al. \cite{10.1145/3510003.3510221} compared energy consumption and run-time performance between PyTorch and TensorFlow, revealing significant differences. Additionally, Fernandes et al. \cite{fernandes2023energy} emphasize the importance of understanding and optimizing software energy consumption. 

Järvenpää et al. \cite{jarvenpaa2023synthesis} have recently proposed 30 \textbf{green architectural tactics} for ML enabled systems, build on top of the study by Verdecchia et al. \cite{verdecchia2023systematic} organized into six categories: model optimization, data-centric, algorithm design, model training, deployment, and management. However, there is currently no evidence on the extent to which energy consumption can be reduced: "more research is required to evaluate the effectiveness of several green architectural tactics in practice" \cite{jarvenpaa2023synthesis}. 

In particular, there is an increasing interest by practitioners in one of these tactics, namely \textbf{model optimization} tactics, evidenced by dedicated libraries by major players (PyTorch, TensorFlow) \cite{pytorchQuantization, tensorflowTensorFlowModel}, and tutorials by the community in Hugging Face \cite{huggingfaceIntroductionModelOptimization}. Model optimization is the process of refining an ML model to reduce computational load and memory usage \cite{lamberti}. It can improve model efficiency and performance, which is vital for energy consumption as well as for other non-functional requirements (a.k.a. quality attributes) like resource constraints, latency requirements, power consumption, and hardware compatibility.

Therefore, software engineers and ML engineers face one challenge: how to integrate the existing \textbf{pre-trained ML models} within \textbf{ML-enabled systems}, considering key quality attributes (e.g., GPU utilization, power and energy consumption, accuracy, time, computational complexity, and economic costs). These pre-trained ML models end up in ML components as specific parts of ML systems \cite{10.1145/3487043}.

In this context, this study is motivated by the observation of a gap in the current research, where limited attention is given to measuring energy consumption and the economic cost in the inference phase of ML models \cite{demartino2024classificationchallengesautomatedapproaches}. With the increasing number of ML models in production and their scalable use, their inference is as important as their training \cite{luccioni2023power} \cite{desislavov2023trends}. Furthermore, in general, ML projects prioritize accuracy metrics, often overlooking the importance of selecting energy-efficient models \cite{10595070}.

The primary objective of this research is to analyze model optimization techniques from the perspective of several quality attributes. We place a strong emphasis on enhancing the energy consumption of pre-trained ML models during the inference stage by means of applying model optimization strategies. We conduct a controlled experiment to systematically evaluate the impact of various optimization techniques from PyTorch (namely, dynamic quantization, \texttt{torch.compile}, local pruning, and global pruning) on the inference quality of ML components, which includes GPU utilization, power and energy consumption, accuracy, time, computational complexity, and economic costs. 

This research has four primary contributions:
 
\begin{enumerate}
    \item \textbf{Analyzing datasets and pre-trained ML models}: We examined Hugging Face's \cite{huggingfaceHuggingFace} top image classification datasets, with ImageNet-1k and CIFAR-10 being the most popular based on likes and downloads. We conducted a stratified sampling of the models trained on popular datasets, noting that many are fine-tuned versions of existing models. 

    \item \textbf{Quantifying the effect of PyTorch ML optimization in inference quality attributes}: Dynamic quantization significantly reduces energy consumption and inference time while maintaining high accuracy. In contrast, \texttt{torch.compile} balances accuracy and energy consumption, and pruning techniques generally lower accuracy, with local pruning being less favorable.

    \item \textbf{Investigating the return of investment and cost reduction with PyTorch ML optimization}: Although dynamic quantization requires more initial optimization effort, it proves to be cost-effective during inference. In contrast, global pruning significantly impacts costs due to longer optimization times.

    \item \textbf{Elaborating guides for PyTorch ML optimization based on prioritized quality attributes}: We recommend dynamic quantization for speed, and no optimization when accuracy and energy consumption are priorities. If accuracy is not crucial and economic implications are minimal, global pruning with the highest pruning amount is recommended. When the trade-off between accuracy and energy consumption is a concern, global pruning with a moderate amount is recommended; otherwise, local pruning is suggested.
\end{enumerate}

\textbf{Data availability statement}: All research components are publicly available on Zenodo \cite{gonzalez2024}, including models, datasets, and the complete Python code (data downloading, preprocessing, inference, and analysis).

\section{Background}
In this section, we cover essential concepts for understanding this study and examine related work in the domain.

\subsection{AI and Sustainability: ML Optimization Tactics}\label{subsec: AI and Sustainability}
When referring to \textbf{inference}, one means to deploy a trained model to make predictions on new data. An Amazon Web Services (AWS) publication from 2022 \cite{aws_2022} highlighted a crucial insight: a substantial 90\% of the overall cost associated with ML is attributed to the inference phase. 

Discussions about AI have expanded to encompass \textbf{Green AI}, aimed at mitigating the environmental footprint of technology through sustainable practices. Cruz et al. \cite{cruz2024innovating} categorize Green AI into three dimensions: \textit{data-centric}, focused on optimizing data preparation; \textit{model-centric}, aiming to develop and optimize ML models that achieve comparable results with fewer resources; and \textit{system-centric}, addressing the optimization of software architecture and deployment practices to enhance environmental sustainability. 

Listed below are some model-centric optimization strategies, focused on optimizing ML models to achieve comparable results with reduced resource consumption. While we will not apply all, they offer a comprehensive overview.

\textbf{Pruning} \cite{pytorchPruningTutorial}: Zeros out a percentage of weights for a sparse neural network representation. Several criteria can be used to decide the prunable parameters: \textbf{random pruning}; \textbf{weight-magnitude pruning}, which removes parameters with the least weight (e.g., L1 norm or L2 norm); or \textbf{gradient pruning}, based on the accumulated gradient and requiring a backward pass. If connections are removed within each layer, it is considered \textbf{local pruning}. By contrast, \textbf{global or grouped pruning} involves selecting and removing a fraction of parameters across all layers. This optimization can eliminate neurons and their connections from each layer individually (\textbf{structured pruning}), or preserve the neural network's shape by removing connections between neurons in adjacent layers (\textbf{unstructured pruning}).
    
\textbf{Quantization} \cite{pytorchQuantization}: Converts model weights from high to lower precision (e.g., 32-bit floating-point to 8-bit integers). For instance, PyTorch offers three types of quantization: \textbf{Dynamic Quantization (DQ)} quantizes weights, with activations read and stored in floating point and quantized for computing, supporting only \textit{nn.Linear} and \textit{nn.LSTM} layers; \textbf{Post-Training Static Quantization (PTSQ)} quantizes weights and activations with post-training calibration; and \textbf{Quantization-aware training (QAT)} models quantization numerics during training.

\textbf{Torch.compile} \cite{pytorchIntroductionTorchcompile}\cite{huggingfaceOptimizeInference}: Introduced in PyTorch 2.0 and officially released in March 2023, \texttt{torch.compile} applies optimizations such as operator fusion and dead code elimination to reduce overhead and GPU read/writes, potentially speeding up models. The effectiveness depends on factors like model complexity and batch size. 
    
\textbf{Knowledge Distillation} \cite{pytorchKnowledgeDistillation}: Introduced by Hinton et al. \cite{hinton2015distilling}, it transfers knowledge from a computationally expensive model (teacher) to a smaller one (student) while maintaining accuracy. The latter becomes cheaper to evaluate and can be deployed on less powerful hardware, making it ideal for small or medium-sized datasets since fewer architecture assumptions are required \cite{goel2020survey}.
 
\textbf{Convolutional Filter Compression and Matrix Factorization} \cite{goel2020survey}: Techniques like SqueezeNet and MobileNet are notable examples of convolutional filter compression. SqueezeNet reduces the number of parameters by converting 3x3 convolutions into 1x1 convolutions, while MobileNet uses depthwise separable convolutions and bottleneck layers to minimize computation, latency, and number of parameters. In the realm of matrix factorization, multidimensional tensors are approximated using smaller matrices. 

Despite extensive research on ML model optimization, most focus on training rather than inference. This study optimizes pre-trained PyTorch models from Hugging Face to enhance inference performance. We investigate the impacts of various optimization techniques on GPU utilization, power and energy consumption, accuracy, time, computational complexity, and economic costs.

\subsection{Quality Attributes Affected by ML Optimization}
The \textbf{Green Software Measurement Model (GSMM)} \cite{GULDNER2024402} offers a structured approach for measuring efficiency. It defines the \textit{measured object} (software), \textit{measurement goals} (e.g., comparing cycles, configurations), \textit{metrics} (e.g., energy consumption, hardware usage), which define the \textit{measurement procedure}, and \textit{measurement setup}. These components contribute to \textit{data evaluation} for generating insightful reports.

A crucial quality attribute is \textbf{inference time}, the duration of each inference, and \textbf{optimization time}, the time needed to optimize each model. Additionally, \textbf{accuracy} is measured by comparing the predicted class to the ground truth.

Enhancing accuracy requires addressing \textbf{energy efficiency} challenges \cite{hampau2022empirical}. Google's 2022 insights \cite{google_2022} estimate that inference comprises about 60\% of their total ML energy consumption. Luccioni et al. \cite{luccioni2023power} rank image classification as the top Computer Vision task by model emissions. The study introduces tools for measuring energy and carbon emissions during inference (e.g., Code Carbon \cite{codecarbonCodeCarbonio} and MLCO2), noting that generative tasks as more energy-intensive than classification and text-based tasks.

In measuring the \textbf{energy consumption} of ML models, it is paramount to consider various metrics \cite{cruzbekker2023energyunits}. \textbf{Energy} denotes the capacity to do work, measured in joules (J) or kilowatt-hours (kWh) (1kWh=3,600,000J). \textbf{Power}, measured in watts (W), indicates energy consumption per second.

Lastly, the \textbf{economic implications} of model optimization involve estimating associated with computer memory usage during both the optimization and inference phases. The \textbf{Software Carbon Intensity (SCI) Specification} \cite{greensoftwareSoftwareCarbon} provides a methodology for calculating the carbon emissions of software applications relative to a functional unit, which in our context is the number of inferences made. The \textbf{Return On Investment (ROI)} \cite{1293070} \cite{wikipediaReturnInvestment} is defined as the ratio of profit to investment, where profit is the difference between benefits and costs.

\section{Experimental Design}
We designed a technology-oriented experiment following well-established methodologies \cite{wohlin2012experimentation}. In this section, we report and describe the most important details associated with this experiment \cite{jedlitschka2008reporting}, namely our experiment objects (ML models, ML optimization techniques, datasets), experiment variables, and experiment execution.

\subsection{Research Objectives}\label{subsec:research_questions}

Following the Goal Question Metric (GQM) template \cite{caldiera1994goal}, our goal is to \textbf{analyze model optimization techniques with the purpose of assessing their impact with respect to GPU utilization, power and energy consumption, accuracy, time, computational complexity, and economic costs from the point of view of an ML engineer in the context of image classification tasks}.

Regarding the context of this research, we focus on a specific ML task, namely \textbf{image classification}. This task involves categorizing an image into one of predefined classes. The choice is justified by its widespread use in research \cite{desislavov2023trends} \cite{luccioni2023power} and its significance in practical applications, ranging from medical imaging to facial recognition. Furthermore, the \textbf{Stanford AI Index Report} \cite{stanfordIndexReport} highlights that AI has surpassed human performance in image classification, emphasizing it as a mature and impactful application of AI. 

We use the \textbf{PyTorch} library, specifically version 2.2.1, for ML optimization due to its extensive adoption and versatility (see Section \ref{subsec:related_work}). Our study explores four optimization techniques within PyTorch: dynamic quantization, \texttt{torch.compile}, local pruning, and global pruning. These techniques are evaluated in comparison to the original model to assess their effects on quality attributes.

Below are the Research Questions (RQs) of the study:

\begin{itemize}
    \item \textbf{RQ1}: How do model optimization techniques, specifically dynamic quantization, pruning, and \texttt{torch.compile} affect quality attributes?

    \begin{itemize}
        \item \textbf{RQ1.1}: What is the average impact of model optimization on GPU utilization, power and energy consumption, accuracy, time, and computational complexity?
    
        \item \textbf{RQ1.2}: How do model optimization techniques affect selected performance metrics?
    \end{itemize}

    \item \textbf{RQ2}: What are the time-related effects of model optimization and the associated economic costs?
\end{itemize}

\subsection{Experimental units and Dataset Construction}\label{sec:ds_construction}

\begin{figure}[h]
    \centering
    \includegraphics[width=1\linewidth]{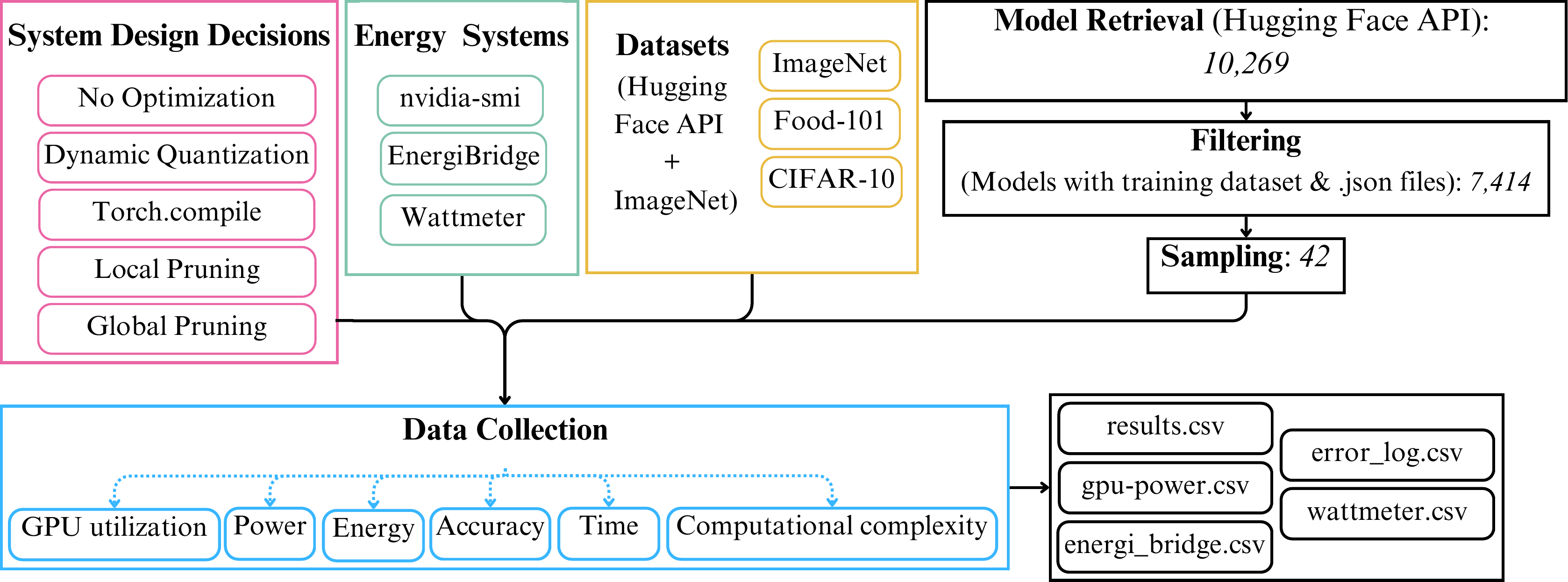}
    \captionsetup{justification=centering}
    \caption{Dataset construction and experiment execution.}
    \label{fig:study_design}
\end{figure}

Building on previous RQs, Figure \ref{fig:study_design} depicts the steps followed to gather and prepare the data for analysis. 

We executed a pipeline using the Hugging Face Hub API \cite{huggingfaceHfApiClient} to collect information on all models uploaded to the platform until March 3rd, 2024. The metrics include model size, training datasets, download and like counts, and the library used. We derived each model's popularity by summing the normalized number of likes and downloads.

The total number of models was 10,269. Given our focus on inference, it was mandatory to employ models with their provided training dataset to ensure testing on images aligned with their training data. Access to all necessary files was crucial, and 7,414 models met these requirements. Most models were from the \textit{transformers} library, with three using the \textit{timm} library and one the \textit{fastai} library. Among the available datasets, we prioritized those with the highest number of models trained on them: 

\begin{itemize}
    \item \textbf{ImageNet} \cite{imagenetImageNet} \cite{deng2009imagenet}: Comprises over 14 million annotated images according to the WordNet hierarchy, with diverse shapes (e.g., 500x375, 800x600...). It is used for the ImageNet Large Scale Visual Recognition Challenge (ILSVRC), with test images and labels disclosed only in 2010. Thus, we use images from ILSVRC2010 whose labels align with the ones from ILSVRC2012, the dataset utilized for model training. 

    \item \textbf{Food-101} \cite{bossard14}: Consists of 101 food categories (750 training and 250 test images for each). All images are represented with 32x32 RBG pixels.

    \item \textbf{CIFAR-10} \cite{krizhevsky2014cifar} \cite{krizhevsky2009learning}: Contains 5,000 training and 1,000 test images for each of the 10 classes. Images are RGB and have varying shapes.
\end{itemize}

Lastly, models trained on the previous datasets were stratified into four quartiles based on their size and popularity. Using stratified sampling, as described by Thompson \cite{thompson2012sampling} and applied in the context of Hugging Face models by Castaño et al. \cite{castano2024lessons}, we selected a representative sample of 42 models (see Section \ref{sec:results-sampling}).

\subsection{Variables}
The variables of the experimental design are grouped into two categories, as outlined in Table \ref{tab:variables}. 

The optimization strategies involved in this study represent the \textbf{independent variables}. The control group for this experiment is represented by the absence of optimization measures, referred to as 'no optimization'. Out of the strategies detailed in Section \ref{subsec: AI and Sustainability}, our experimental groups encompass dynamic quantization, \texttt{torch.compile}, and pruning, chosen for their potential to enhance computational efficiency and reduce energy consumption, with quantization and pruning being the most commonly used techniques \cite{huggingfaceOptimumHardware}. Regarding dynamic quantization, \textit{Linear} and \textit{Long Short-Term Memory (LSTM)} layers are quantized from 32-bit floating point to 8-bit integer. For the case of pruning, the experiments include both, local structured pruning and global unstructured pruning with varying pruning amounts, using magnitude-based pruning (L1 norm). This means that weights are ranked based on their magnitudes, and the ones that lie below a threshold are set to 0. The following layers are targeted for optimization: \textit{torch.nn.Linear}, \textit{torch.nn.Conv1d}, \textit{torch.nn.Conv2d}, \textit{torch.nn.Conv3d}, \textit{torch.nn.LayerNorm}, \textit{torch.nn.ConvTrans pose1d}, \textit{torch.nn.ConvTranspose2d}, \textit{torch.nn.ConvTranspos e3d}, \textit{torch.nn.Embedding}, \textit{torch.nn.MultiheadAttention}, \textit{torc h.nn.BatchNorm2d}, and \textit{torch.nn.InstanceNorm2d}. It is important to note that PyTorch pruning does not explicitly remove pruned parameters. Instead, it masks them to zero, resulting in no decrease when assessing metrics such as model size, FLOPs, MACs, and the number of parameters. However, the model gets more sparse when increasing the pruning amount. 

The \textbf{dependent variables} include computational metrics such as GPU utilization, along with power and energy consumption (both for individual components like the CPU and GPU, and for global consumption). The accuracy of the models is collected, as well as the optimization and inference times. The study further examines the costs associated with model optimization and inference. Computational complexity is determined by the resources needed to execute an algorithm for a given input. \textit{Model parameters} directly influence this, with larger models typically requiring more resources. The \textit{number of operations}, such as Floating Point Operations (FLOPs) and Multiply-Accumulate Operations (MACs) are key metrics. It is widely accepted that one MAC is equivalent to two FLOPs \cite{pytorch-opcounter}. Desislavov et al. \cite{desislavov2023trends} highlight an exponential growth in Deep Learning (DL) model parameters, noting that parameter count does not directly correlate with computational requirements, particularly in Convolutional Neural Networks (CNNs) where different distributions can lead to varied FLOP counts. Additionally, the study evaluates several cost-related metrics: Mean Optimization Cost (MOC) reflects the average cost incurred during model optimization; Mean Inference Cost (MIC) represents the average cost for the inference stage; Cost Per Inference (CPI) calculates the financial cost associated with each individual inference; Cost Savings Per Inference (CSPI) assesses the cost reduction achieved per inference due to optimization strategies; and Number of Inferences (NI) denotes the total number of inferences performed throughout the experiment.

\begin{table}[ht]
    \caption{Description and type of independent and dependent variables in the study.}
    \label{tab:variables}
        \begin{tabular}{ccc}
         \toprule
         \multicolumn{3}{c}{\textbf{Independent Variables}} \\
         
         \midrule
         \textbf{Variable} & \textbf{Description} & \textbf{Type} \\
         \midrule
         Optimization & Optimization strategy & Nominal \\

         \toprule
         \multicolumn{3}{c}{\textbf{Dependent Variables}} \\
         
         \midrule
         \textbf{Variable} & \textbf{Description} & \textbf{Type} \\
         \midrule
         GPU Utilization & Utilization of computational hardware & Numerical \\
         \midrule
         Power & Power consumption (CPU, GPU, or overall server) & Numerical \\
         \midrule
         Energy & Energy consumption (CPU, GPU, or overall server)  & Numerical \\
         \midrule
         Accuracy  & Correct Predictions/Total Predictions & Numerical \\
          \midrule
         Optimization Time & Time needed to optimize a model& Numerical \\
         \midrule
         Inference Time & Time needed to run each inference & Numerical \\
         \midrule
         Computational Complexity& FLOPs, MACs, and number of parameters & Numerical \\
         \midrule
         MOC  & Mean Optimization Cost & Numerical \\
         \midrule
         MIC & Mean Inference Cost & Numerical \\
         \midrule
         CPI & Cost Per Inference & Numerical \\
         \midrule
         CSPI & Cost Savings Per Inference & Numerical \\
         \midrule
         NI & Number of Inferences & Numerical \\
         \bottomrule
    \end{tabular}
\end{table}

Furthermore, \textbf{other variables} not relevant for the RQs are considered and controlled, such as the measuring instrument, model size, file size quartile, popularity quartile, training dataset, and memory usage.

\subsection{Experiment Execution} 

For each of the selected models and its optimized version, we conduct 100 inferences (NI) using different test images. It is crucial to repeat the experiments several times to guarantee that the obtained results are reliable. An accepted number of measurements among researchers is 10 \cite{luccioni2023power}. The experiments are executed on our high-performance server, equipped with an NVIDIA RTX 4090 24GB GPU, AMD Ryzen 9 7950X CPU with 16 cores and 32 threads clocked at 4.5 GHz, 32GB of DDR5 5600 MHz RAM. The pipeline is executed over six days at the end of April 2024. 

It is worth mentioning that is not possible to parallelize the experiments, in the sense of running simultaneous inferences for different models. Accurate measurements require isolating each inference, as parallel execution could interfere with resource usage and affect reliability.

Regarding energy consumption, it is sourced from three systems: 
    \begin{enumerate}
        \item \textbf{NVIDIA System Management Interface (nvidia-smi)} \cite{nvidia-smi}: A command-line utility providing information about GPU utilization, memory usage, and power draw. The provided data is logged in a \textit{.csv} file every 100 milliseconds. 

        \item \textbf{EnergiBridge} \cite{sallou2023energibridge}: An energy measurement utility to collect CPU resource usage data. This information is saved in a \textit{.csv} file every 200 milliseconds. The energy consumption recorded by this tool is accumulated over time and not just for each instance.
        
        \item \textbf{Wattmeter} \cite{netioproducts}: A device for measuring the power consumption of a system every 500 milliseconds in a \textit{.csv} file. Energy metrics here represent the total consumption since the last reset of the counter. 
    \end{enumerate}

The output of this stage consists of five \textit{.csv} files named after the corresponding task to which the models belong and the dataset used for training them.

\subsection{Energy Data Quality Check and  Normalization}
We use the timestamps to map accuracy and inference duration with energy metrics from the monitoring systems. Some inferences may correspond to multiple entries in each file, in which case the values are averaged.

Following this, we validate that the predictions are as expected. Specifically, in case any model has predicted a category name instead of a category number, we replace the prediction with the expected value. Subsequently, we concatenate the data into a unified \textit{.csv} file.

Next, we verify consistency by ensuring that the sum of CPU energy (from EnergiBridge) and GPU energy (from nvidia-smi) is lower than the global energy recorded by the wattmeter. This process implies converting GPU power (in watts) to energy values (in joules) by multiplying by time. 

To facilitate analysis and comparison across optimizations, we combine accuracy and energy into a new metric proposed by Hanafy et al. \cite{hanafy2021design} to capture the trade-off. It quantifies the energy required per unit of accuracy achieved.

\subsection{Data Analysis and Hypothesis}\label{subsec:data_analysis}
We analyze the collected data for answering the RQs. 

\subsubsection{RQ1 Analysis}
To study the average impact of optimizations on the dependent variables (RQ1.1), we use a structured approach. Initially, we group the data according to each model, optimization, and repetition, and calculate the mean value for each variable. Following this, we aggregate the mean values associated with each optimization. 

We propose an underlying relationship between dependent variables and the optimization strategy used, aiming to analyze the impact of each optimization on these variables (RQ1.2). A linear model is utilized to identify which optimizations most significantly affect the dependent variables. The method of least squares is applied to fit models for inference time, CPU energy, and global load (which includes all power consumed by the computer), as follows:

$$Inference \ Time = \beta_1 Optimization + \epsilon_1$$
$$CPU \ Energy = \beta_2 Optimization + \epsilon_2$$
$$Global \ Load = \beta_3 Optimization + \epsilon_3$$

where \( Optimization \) is a column representing all the different strategies; \( \beta_i \) is the coefficient indicating the impact of the optimizations on the i-th dependent variable; and \( \epsilon \) is the error term, which is expected to have low magnitude. 

The first inference for each model and repetition has been discarded from this analysis and will be discussed in detail in RQ2 and Section \ref{subsec:IMG_economic}.

\subsubsection{RQ2 Analysis}
To assess the impact of model optimization, we start by checking if the first inference might be influenced by compilation steps that are not representative of the performance during later inferences. We conduct hypothesis testing by comparing the time of the first inference with the mean of subsequent ones, as summarized in Table \ref{tab:hypothesis}. We first perform a Shapiro-Wilk test for normality, with a significance level of $\alpha = 0.05$. If the normality assumption is not violated (p-value \(>\) 0.05), we use a parametric test for dependent samples: t-test. Otherwise, the non-parametric test is Wilcoxon signed-rank test, which compares two dependent samples without imposing normality. 

$$
\left\{
\begin{array}{ll}
H_0: & \mu_d = 0 \\
H_1: & \mu_d \neq 0
\end{array}
\right.
$$

Here, \( \mu_d \) represents the population mean difference between the paired observations. The null hypothesis ($H_0$) assumes no difference exists, while the alternative hypothesis ($H_1$) suggests a difference does exist.

We incorporate a power analysis to determine the study's ability to detect true effects based on sample size and observed effect size. If the power is below the threshold of 0.8, it indicates a potential risk of Type II error, which occurs when a study fails to reject a null hypothesis that is false. Additionally, we calculate Cohen's d, a standardized effect size measuring two group mean differences \cite{cohen2013statistical}. 

\begin{table}[ht]
    \caption{Statistical tests performed for RQ2.}
    \label{tab:hypothesis}
    \begin{tabular}{c p{5.5cm} cc}
         \toprule
         \textbf{Variables} & \centering \textbf{Null Hypothesis} & \textbf{Statistical Test} & \textbf{Results} \\
         \midrule
         \multirow{2}{8em}{Initial vs. mean subsequent inference time} & The data is normally distributed & Shapiro-Wilk & Table \ref{tab:img_wilcox} \\
         \cline{2-4}
          & There is no difference between these pairs in the population & Wilcoxon & \\
         \bottomrule
    \end{tabular}
\end{table}

Next, we examine the time and the economic implications of the optimization process, encompassing model loading, optimization, and the first inference. To incorporate economic considerations, we use the memory reported by EnergiBridge, which details memory consumption across the system. Cost estimates are based on packs, each allowing the usage of 1 core of CPU for 1 hour and 4 gigabytes for €0.4 plus taxes. This approach is derived from observed costs across cloud services, such as the EC2 service located in Spain \cite{amazonProductService}, providing us with a cost proxy for relative comparisons. The analysis involves calculating the maximum amount of packs needed per optimization. This economic study extends to the operational part, which encompasses the inferences. For ROI, we assess the ratio of net financial gain to the total investment, calculated using the formula:

$$\text{ROI} = \frac{\text{CSPI} \cdot \text{NI} - (\text{MOC} + \text{CPI} \cdot \text{NI})}{\text{MOC}+\text{CPI} \cdot \text{NI}}$$

Lastly, we calculate the break-even point, which is the point at which the total costs equal the benefits \cite{felderer2020contemporary}.

\section{Results}
In this section, we first provide context on our experimental units (studied datasets and pre-trained models). Then, we reply to the RQs.

\subsection{Analysis of Image Classification Datasets and Models in Hugging Face}\label{sec:results-sampling}

\subsubsection{What are the most popular datasets?}
\begin{figure}[!ht]
    \centering
    \includegraphics[width=1\linewidth]{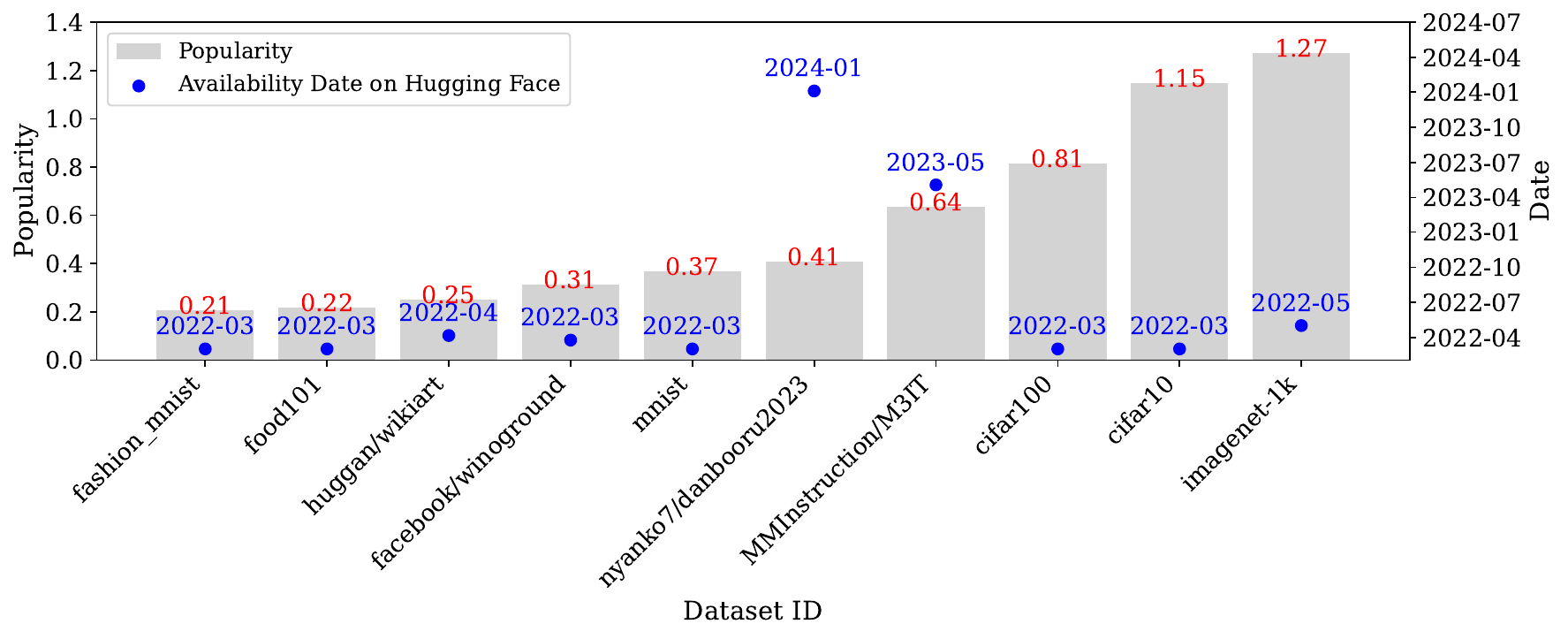}
    \captionsetup{justification=centering}
    \caption{Top 10 most popular image classification datasets.}
    \label{fig:IMG_top10_ds}
\end{figure}

Figure \ref{fig:IMG_top10_ds} shows the ten most popular image classification datasets on Hugging Face, determined by the sum of their normalized likes and downloads. The bars in the chart represent the level of popularity of each dataset, and the points indicate the dates when they were uploaded to the platform. Notably, two datasets emerge as the most popular: ImageNet-1k and CIFAR-10, which have been foundational in the field of Computer Vision, serving as benchmarks for evaluating the performance of new models. 

Table \ref{tab:IMG_top10_tasks} shows the publication dates and supported ML tasks for the ten most popular datasets. Some datasets are older, having been published years before they were posted on Hugging Face. The presence of older datasets indicates that they have gained significant importance and continue to be highly utilized, possibly due to their wide applicability in various ML applications. It is also worth mentioning that this data was collected in early 2024, so newer datasets have not had enough time to achieve the same level of recognition and usage as their predecessors. While Image Classification (IC) is predominant, others like Image-to-Text (ItT) and Text-to-Image (TtI) are also covered. Some datasets include Multi-class Image Classification (MIC) challenges, showcasing their versatility.

\begin{table}[ht]
\caption{Top 10 image classification datasets with publication dates and associated tasks (\checkmark) and sub-tasks (*).}
\label{tab:IMG_top10_tasks}
\begin{tabular}{@{}llcccc@{}}
\toprule
\multirow{2}{*}{\textbf{Dataset}} & \multirow{2}{*}{\textbf{Publication}} & \multicolumn{4}{c}{\multirow{2}{*}{\textbf{Tasks}}} \\ 
 & \multirow{2}{*}{\textbf{Date}} & \multicolumn{4}{c}{} \\
\cmidrule(lr){3-6} 
 & & IC & ItT & TtI & MIC \\
\midrule
         fashion\_mnist & 2017 \cite{xiao2017fashion} & \checkmark & & & * \\
        \hline
         food101 & 2014 \cite{bossard2014food} & \checkmark & & & * \\
        \hline
         huggan/wikiart & 2022 \cite{huggingfaceHugganwikiartDatasets} & \checkmark &  \checkmark & \checkmark & \\
         \hline
         facebook/winoground & 2022 \cite{thrush2022winogroundprobingvisionlanguage} & \checkmark &  \checkmark & \checkmark & \\
        \hline
         mnist & 1998 \cite{lecunMNISTHandwritten} & \checkmark & & & * \\
         \hline
         nyanko7/danbooru2023 & 2024 \cite{huggingfaceNyanko7danbooru2023Datasets} & \checkmark & \checkmark & \checkmark & \\
         \hline
         MMInstruction/M3IT & 2023 \cite{li2023m3it} & \checkmark & \checkmark &  & \\
         \hline
         cifar100 & 2009 \cite{krizhevsky2014cifar} & \checkmark &  & & \\
         \hline
         cifar10 & 2009 \cite{krizhevsky2014cifar} & \checkmark &  & & \\
         \hline
         imagenet-1k & 2009 \cite{deng2009imagenet} & \checkmark & & & * \\
\bottomrule
\end{tabular}
\end{table}

\subsubsection{What insights can be derived from analyzing the metadata of the models?}

Let us use image classification models with the necessary files for inference and whose training data is available. Notably, ImageNet, Food-101, and CIFAR-10 are the three most used training datasets, with 156, 123, and 49 models, respectively. Other datasets appear in only a few models.

\begin{figure}[!ht]
    \centering
    \includegraphics[width=0.9\linewidth]{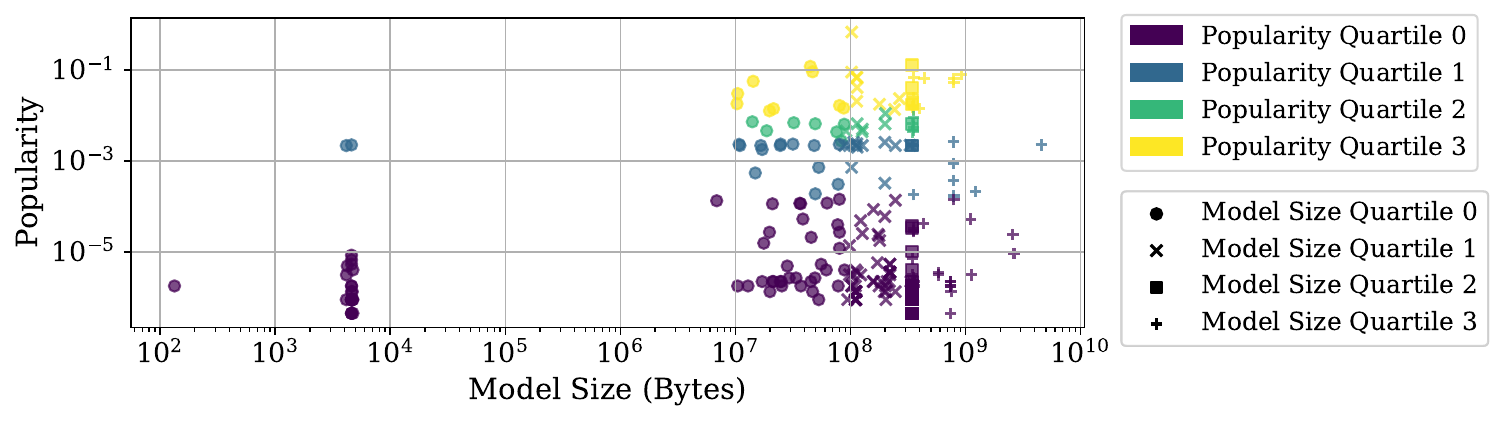}
    \captionsetup{justification=centering} 
    \caption{Relationship between model size and popularity.}
    \label{fig:IMG_quartiles}
\end{figure}

The relationship between model size and popularity for those models can be seen in Figure \ref{fig:IMG_quartiles}. The vast majority of models are unpopular (popularity quartile 0). Model sizes vary widely, spanning several orders of magnitude.

Regarding the 42 sampled models, each of them underwent individual inspection to check the details given by the author and to ensure that it had not been previously optimized. Notably, most of the models revealed to be fine-tuned versions of existing ones. Consequently, the analysis focuses not on comparing different models within a family (such as ResNet) but rather on studying the performance of these variants. Details on the base models and the number of models fine-tuned from each can be seen in Table \ref{tab:fine-tunings}. These models encompass various architectures, including the Vision Transformer (ViT), and ConvNeXT (model composed solely of Convolutional Neural Network (ConvNet) modules). Some other models fall under the "not specified" category, including base models published by prominent entities like Facebook, Nvidia, or SenseTime. 

\begin{table}[ht]
    \caption{Fine-tuned (FT) models for image classification.}
    \label{tab:fine-tunings}
    \begin{tabular}{c c c} 
        \toprule
        \textbf{Base Model} & \textbf{Architecture} & \textbf{FT} \\
        \midrule
        google/vit-base-patch16-224-in21k & ViT & 20 \\
        \midrule 
        Not Specified & - & 9 \\
        \midrule
        microsoft/swin-base-patch4-window7-224 & ViT & 6 \\
        \midrule
        microsoft/swin-tiny-patch4-window7-224 & ViT & 6 \\
        \midrule
        facebook/convnext-tiny-224 & ConvNeXT & 1 \\
        \bottomrule
    \end{tabular}
\end{table}

All models accept a 3x224x224 input tensor, where 3 represents the RGB channels and 224x224 is the image size.

\bigskip
\noindent
\fcolorbox{black}{white}{
    \parbox{0.95\linewidth}{
    \textbf{Datasets}: Selected based on their popularity and training usage: ImageNet, Food-101, and CIFAR-10.
    
    \textbf{Pre-trained Models}: Emphasizing representative models trained on those datasets, especially fine-tuned ones.
    }
}

\subsection{How do model optimization techniques, specifically dynamic quantization, pruning, and \texttt{torch.compile} affect quality attributes? (RQ1)}

\subsubsection{What is the average impact of model optimization on GPU utilization, power and energy consumption, accuracy, time, and computational complexity?}

\begin{figure}[h]
    \centering
    \includegraphics[width=1\linewidth]{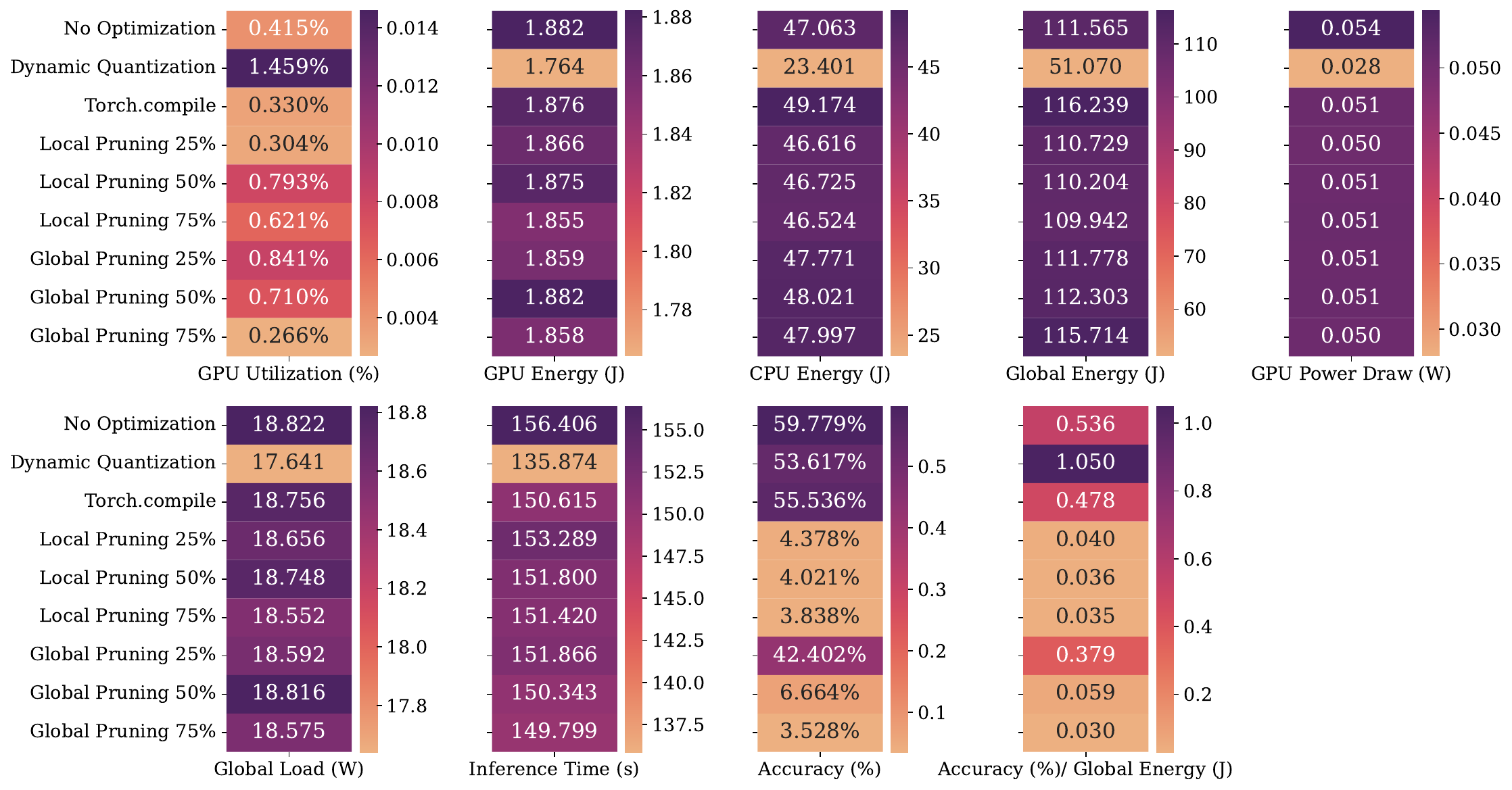}
    \captionsetup{justification=centering}
    \caption{Average performance metrics across optimizations.}
    \label{fig:IMG_heatmaps}
\end{figure}

Figure \ref{fig:IMG_heatmaps} shows heatmaps where darker colors indicate higher mean values.

\textbf{No Optimization}: The baseline scenario demands the highest power consumption. Global load values for the baseline are around 150W, which is equivalent to charging a smartphone for 30 hours \cite{energysageManyWatts}. While the baseline maintains high accuracy, its energy inefficiency and slower inference time highlight the need for optimization.

\textbf{Dynamic Quantization}: This technique is the most resource-intensive in terms of GPU usage but manages to consume relatively low power. It significantly reduces energy consumption to less than half of other techniques and remarkably enhances inference speed, achieving nearly twice the speed compared to others. In terms of accuracy, dynamic quantization performs well, making it a dominant optimization technique overall.

\textbf{Torch.compile}: This optimization technique consumes slightly more energy, but still performs efficiently. It closely approximates the baseline in terms of accuracy, making it a good alternative for those seeking a balance between performance and power consumption.

\textbf{Global Pruning}: At a 25\% pruning ratio, global pruning demonstrates good accuracy. However, as the pruning amount increases, accuracy drastically decreases due to the model becoming more sparse. Power and energy consumption across pruning ratios remain quite similar.

\textbf{Local Pruning}: It shows a drastic decrease in accuracy across all cases, with higher pruning amounts leading to worse predictions. The increased sparsity significantly impacts the model's performance, making local pruning less favorable compared to other techniques.

Overall, there is a trade-off between accuracy and energy consumption. Dynamic quantization offers the best performance in terms of energy efficiency and inference time while maintaining acceptable accuracy. \texttt{Torch.compile} and global pruning at a 25\% ratio are also notable, providing a reasonable compromise between these factors.

\begin{figure}[h]
    \centering
    \includegraphics[width=1\linewidth]{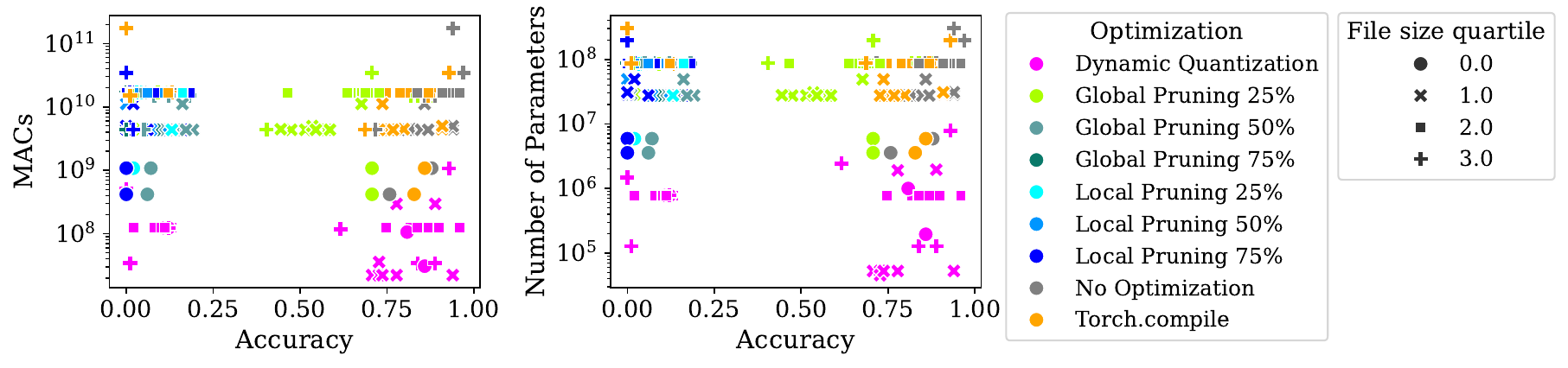}
    \captionsetup{justification=centering}
    \caption{Average computational complexity of optimizations.}
        \label{fig:IMG_macs_params}
\end{figure}

The left plot in Figure \ref{fig:IMG_macs_params} shows the relationship between MACs and accuracy, whereas the right plot depicts the relationship between parameters and accuracy. The y-axis has been logarithmically scaled to facilitate the analysis. Larger models (shape +) have more MACs and parameters. Dynamic quantization is the best configuration, reducing computations while maintaining high accuracy. No optimization and \texttt{torch.compile} follow, maintaining high accuracy without reducing model size. Next is global pruning 25\% with 0.5 accuracy. Lastly, global pruning with higher pruning amounts and local pruning achieve lower accuracy.

\begin{figure}[h]
    \centering
    \includegraphics[width=1\linewidth]{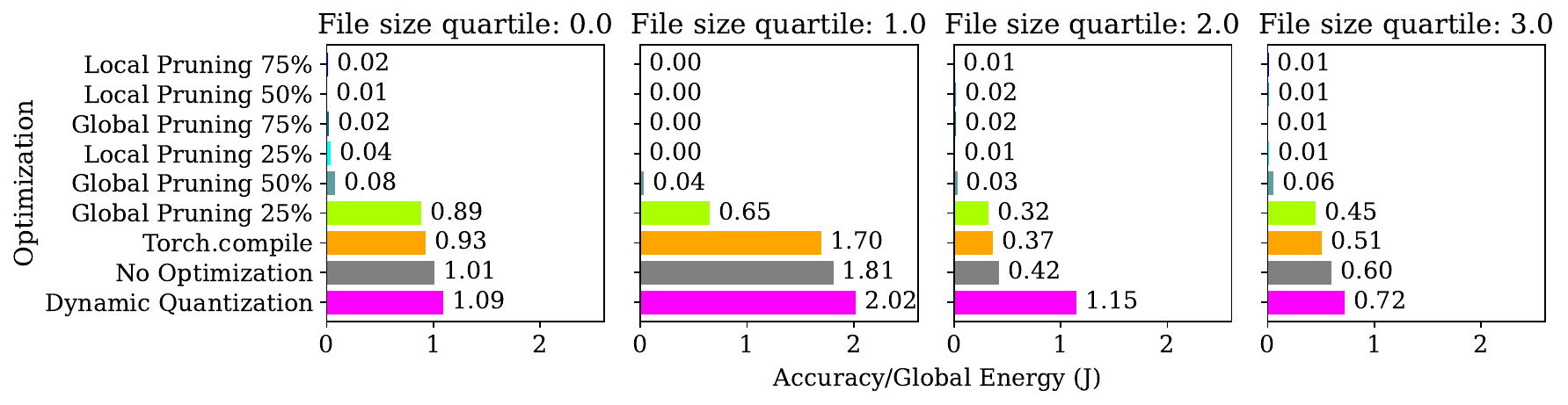}
    \captionsetup{justification=centering}
    \caption{Trade-off between accuracy and energy consumption across model size quartiles.}
    \label{fig:IMG_mdlsize_tradeoff}
\end{figure}

The impact of optimizations on the trade-off between accuracy and energy consumption varies across file size quartiles, as depicted in Figure \ref{fig:IMG_mdlsize_tradeoff}. Global pruning (at 50\% and 75\% at pruning ratios) and local pruning result in negligible accuracy improvements relative to energy consumption across all quartiles. The most effective strategies, ranked in order, are dynamic quantization, no optimization, \texttt{torch.compile}, and global pruning 25\%. For the smallest (quartile 0) and largest models (quartile 3), those strategies achieve similar values. However, the value across all optimizations is higher for the smallest ones. The most significant differences are observed in quartile 1 (small-medium models) and quartile 2 (medium-large models). In the former, global pruning 25\% offers a significantly lower value compared to other optimization techniques. In the latter, dynamic quantization demonstrates clear dominance. 

\begin{figure}[h]
    \centering
    \includegraphics[width=1\linewidth]{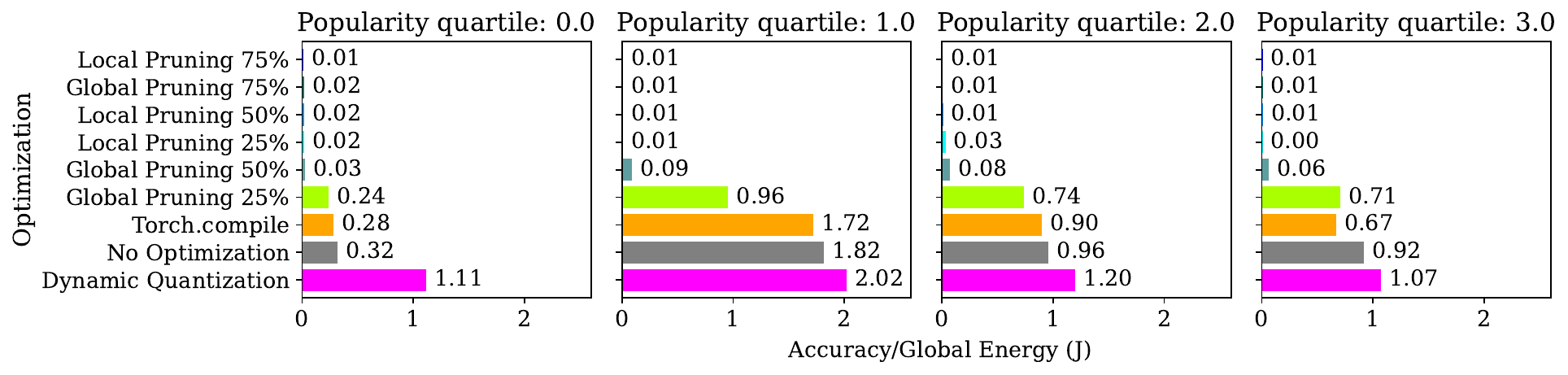}
    \captionsetup{justification=centering}
    \caption{Trade-off between accuracy and energy consumption across popularity quartiles.}
    \label{fig:IMG_pop_tradeoff}
\end{figure}

For popularity quartiles, Figure \ref{fig:IMG_pop_tradeoff} illustrates similar trends to the ones presented for file size quartiles. Dynamic quantization performs best for unpopular models (quartile 0), while for moderately popular models (quartile 1), global pruning 25\% lags behind. Global pruning 25\% outperforms \texttt{torch.compile} in the most popular models (quartile 3).

\noindent
\fcolorbox{black}{white}{
    \parbox{0.95\linewidth}{
    \textbf{Finding 1.1}: Dynamic quantization demonstrates a nearly two-fold increase in speed at the cost of slightly reducing accuracy and consuming more GPU resources.

    \textbf{Finding 1.2}: Torch-compiled models demonstrate a balance between accuracy and energy consumption.
    
    \textbf{Finding 1.3}: Local pruning has no positive effect on accuracy. 
    
    \textbf{Finding 1.4}: Global pruning 25\% can be an alternative for optimizing models although it is less efficient in balancing accuracy and energy consumption compared to dynamic quantization and \texttt{torch.compile}.

    \textbf{Finding 1.5}: The smallest and largest models benefit similarly from top strategies. Dynamic quantization excels for unpopular models.
    }
}

\subsubsection{How do model optimization techniques affect selected performance metrics?}

Table \ref{tab:IMG_time_lm} examines the impact of optimizations on inference time. The estimates represent the time increase or decrease with respect to the baseline. The significance levels used are as follows: high significance ('***'), very significant ('**'), significant ('*'), and minimal significance ('.'). Dynamic quantization significantly reduces time, while \texttt{torch.compile} slightly increases it. Local and global pruning show non-significant decreases, with p-values higher than the significance level (0.05), which does not allow rejecting the null hypothesis that there is no effect. The $R^2$ indicates that 0.2\% of the variability in inference time is explained by the independent variables, suggesting other factors may have more impact on inference time.

\begin{table}[!h]
    \caption{Linear fit between inference time (s) and optimization.}
    \label{tab:IMG_time_lm}
    \begin{tabular}{ccccc}
    \toprule
    \multicolumn{2}{c}{\textbf{Method}} & \textbf{Estimate} & \textbf{P-value} & \textbf{Significance Level}  \\  
    \midrule
    \multicolumn{2}{c}{(Intercept)}  & 0.054462   & \(<\) 2e-16 & *** \\
    \midrule
    \multicolumn{2}{c}{Dynamic Quantization} & -0.026553 & \(<\) 2e-16 & *** \\
    \midrule
    \multicolumn{2}{c}{Torch.compile} &  0.028149  & \(<\) 2e-16 & *** \\
    \midrule
    \multirow{3}{*}{Local Pruning} & 25\% & -0.004023  & 0.0504 & . \\
    & 50\%    & -0.003884  & 0.0588 & . \\
    & 75\%   & -0.003601  & 0.0799 & . \\
    \midrule
    \multirow{3}{*}{Global Pruning} & 25\%  & -0.003795  & 0.0649 & . \\
 & 50\% & -0.003665  & 0.0746 & .   \\
 & 75\% & -0.003996  & 0.0519 & .   \\
     \bottomrule
\end{tabular}
\footnotetext{$R^2 = 0.00196$, Adjusted $R^2 = 0.00193$, F-statistic = 90.42, p \(< 2.2 \times 10^{-16}\)}
\end{table}

Moving forward, Table \ref{tab:IMG_energy_lm}  shows the results for CPU energy consumption. Dynamic quantization significantly increases energy consumption. \texttt{Torch.compile} and global pruning also increase it with a low p-value, indicating statistical significance. For the case of local pruning, lower pruning amounts result in a higher p-value (less statistical significance). The $R^2$ indicates that 1.35\% of the variability in CPU energy consumption is explained.

\begin{table}[!h]
    \caption{Linear fit between CPU energy (J) and optimization.}
    \label{tab:IMG_energy_lm}
    \begin{tabular}{ccccc}
    \toprule
    \multicolumn{2}{c}{\textbf{Method}}& \textbf{Estimate} & \textbf{P-value} & \textbf{Significance Level}  \\  
    \midrule
\multicolumn{2}{c}{(Intercept)} & 20,874,625 & \(<\)  2e-16 & *** \\
    \midrule
\multicolumn{2}{c}{Dynamic Quantization} & 4,192,204 & \(<\)  2e-16 & *** \\
   \midrule
\multicolumn{2}{c}{Torch.compile}  & 450,649 &  1.18e-09 & *** \\
   \midrule
\multirow{3}{*}{Local Pruning} & 25\% & 154,841 &  0.036596 & *   \\
                              & 50\%  & 201,856 &  0.006432 & **  \\
                              & 75\% & 248,923 &  0.000779 & *** \\
  \midrule
\multirow{3}{*}{Global Pruning} & 25\% & 297,394 &  5.96e-05 & *** \\
                               & 50\%  & 347,753 &  2.67e-06 & *** \\
                               & 75\% & 399,100 &  7.15e-08 & *** \\
       \bottomrule
\end{tabular}
\footnotetext{$R^2 = 0.01346$, Adjusted $R^2 = 0.01344$, F-statistic = 629.7, p \(< 2.2 \times 10^{-16}\)}
\end{table}

Lastly, Table \ref{tab:IMG_load_lm} provides insights into the impact of optimizations on global load. Dynamic quantization demonstrates the most substantial decrease. \texttt{Torch.compile} and most pruning cases show significant decreases, as evidenced by their low p-values. However, global pruning 50\% shows no significant decrease in the metrics.

\begin{table}[!h]
    \caption{Linear fit between global load (W) and optimization.}
    \label{tab:IMG_load_lm}
    \begin{tabular}{ccccc}
    \toprule
    \multicolumn{2}{c}{\textbf{Method}}& \textbf{Estimate} & \textbf{P-value} & \textbf{Significance Level}  \\  
    \midrule
\multicolumn{2}{c}{(Intercept)} & 18.820479 & \(<\) 2e-16 & *** \\
    \midrule
\multicolumn{2}{c}{Dynamic Quantization} & -1.179803 & \(<\) 2e-16 & *** \\
    \midrule
\multicolumn{2}{c}{Torch.compile}  & -0.064129 &  0.000104 & *** \\
    \midrule
\multirow{3}{*}{Local Pruning} & 25\% & -0.164418 & \(<\) 2e-16 & *** \\
                              & 50\%  & -0.071172 & 1.66e-05 & *** \\
                              & 75\% & -0.268236 & \(<\) 2e-16 & *** \\
    \midrule
\multirow{3}{*}{Global Pruning} & 25\% & -0.231264 & \(<\) 2e-16 & *** \\
                               & 50\%  & -0.006744 & 0.683215  &    \\
                               & 75\% & -0.247450 & \(<\) 2e-16 & *** \\
       \bottomrule
\end{tabular}
\footnotetext{$R^2 = 0.02053$, Adjusted $R^2 = 0.02051 $, F-statistic = 967.6, p \(< 2.2 \times 10^{-16}\)}
\end{table}

\bigskip
\noindent
\fcolorbox{black}{white}{
    \parbox{0.95\linewidth}{
    \textbf{Finding 1.6}: Model optimizations impact inference time differently, with dynamic quantization notably reducing time and \texttt{torch.compile} showing a slight increase.
    
    \textbf{Finding 1.7}: CPU energy consumption varies significantly with optimizations, with local pruning showing the highest p-values and the least statistical significance.

    \textbf{Finding 1.8}: Optimization methods affect global load diversely: dynamic quantization leads to a substantial decrease. In comparison, \texttt{torch.compile} and pruning result in a moderate reduction.
    }
}

\subsection{What are the time-related effects of model optimization and the associated economic costs? (RQ2)}\label{subsec:IMG_economic}

\begin{figure}[h]
    \centering
    \includegraphics[width=1\linewidth]{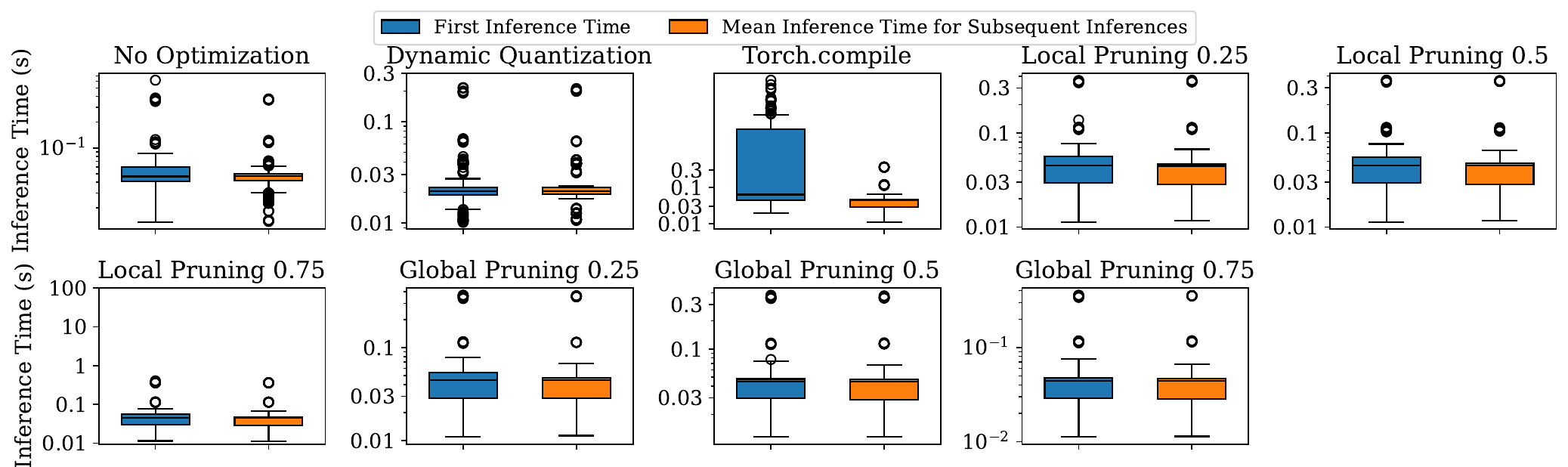}
    \captionsetup{justification=centering}
    \caption{Boxplots of inference times across optimizations.}
    \label{fig:IMG_boxplots_time}
\end{figure}

To comprehensively assess the effects associated with model optimization, it is crucial to first evaluate the initial inference using an optimized model, as it might involve preliminary optimization processes that are not representative of the performance during subsequent inferences. Figure \ref{fig:IMG_boxplots_time} presents an exploratory analysis of the distribution of the time needed to conduct the first inference once the model is optimized, and the mean of the subsequent ones. The y-axis scale has been logarithmically adjusted to accommodate the data's range of values. Despite the existence of outliers, the first inference takes longer, which is particularly evident for \texttt{torch.compile}. The initial execution for \texttt{torch.compile} may take longer due to compilation, but subsequent runs are faster as they use an optimized kernel.

We conduct a statistical analysis, starting with a Shapiro-Wilk test for each of the optimization techniques. Table \ref{tab:img_wilcox} displays the test results. In all cases, the p-values are lower than 0.05, rejecting the null hypothesis of normality. 

Next, we perform a Wilcoxon test to assess whether there is a significant difference between the time taken for the first inference and the average time taken for subsequent inferences for each optimization technique. All optimizations have p-values lower than 0.05, except global pruning with 25\% and 75\% pruning amounts, where there is no significant evidence to reject the null hypothesis of equal differences. 

Additionally, power analysis provides insights into the sensitivity of the Wilcoxon test. The power value for \texttt{torch. compile} is 1, indicating a high likelihood of correctly rejecting the null hypothesis when true differences exist. For other optimizations, the low power suggests the need for larger sample sizes or stronger effect sizes to detect differences in inference times reliably.

\begin{table}[ht]
    \captionsetup{justification=centering} 
    \caption{Shapiro-Wilk and Wilcoxon Test Results.}
    \label{tab:img_wilcox}
    \begin{tabular}{cccccc}
         \toprule
         \multicolumn{2}{c}{\textbf{Optimization}} & \multicolumn{2}{c}{\textbf{Shapiro-Wilk Test}} & \multicolumn{2}{c}{\textbf{Wilcoxon Test}} \\
         \cline{3-6}
         
         \multicolumn{2}{c}{\textbf{Technique}} & \centering \textbf{Test Statistic} & \textbf{P-value} & \centering \textbf{Test Statistic} & \textbf{P-value}\\
         
         \midrule
         \multicolumn{2}{c}{No Optimization} & 0.11 & 5.63e-40 & 49,552 & 6.75e-03 \\
        \midrule
        \multicolumn{2}{c}{Dynamic Quantization} & 0.65 & 7.11e-28 & 33,222 & 2.08e-04 \\
        \midrule
        \multicolumn{2}{c}{Torch.compile} & 0.41  & 3.10e-34  & 74,340 & 4.84e-41\\
        \midrule
        \multirow{3}{*}{Local Pruning} &  25\% & 0.76  & 6.87e-24  & 54,117  & 5.92e-07\\
        & 50\% & 0.56 & 1.37e-30 & 48,372 & 9.30e-03\\
        & 75\% & 0.66  & 1.81e-27  & 52,358 & 2.03e-05\\
        \midrule
        \multirow{3}{*}{Global Pruning} & 25\% & 0.74 & 1.86e-24  & 43,644 & 5.20e-01\\
        & 50\% & 0.72  & 1.56e-25  & 36,187 & 1.33e-02\\
        & 75\% & 0.74  & 1.15e-24  & 41,138 & 6.80e-01\\
        \bottomrule
    \end{tabular}
\end{table}

To ensure consistency, even though significance was found only for \texttt{torch.compile}, the first inference is considered part of the optimization process in all cases.

Table \ref{tab:IMG_cost} shows the mean (avg) and standard deviation (SD) of the time for accessing the model, optimizing it, and conducting the first inference, referred to as the \textbf{total optimization time}. Dynamic quantization takes a slightly longer time than no optimization. \texttt{Torch.compile} has a significantly longer standard deviation, over 4.5 seconds. Global pruning results in much longer times compared to other strategies. This is reflected in the MOC, which is particularly high for \texttt{torch.compile} and global pruning. In the former case, memory usage dominates the costs, while in the latter, the extended time impacts the cost. 

For \textbf{inference costs}, \texttt{torch.compile} is slightly more expensive than the baseline. However, the other strategies are more cost-effective, with dynamic quantization being the cheapest due to reduced model size and faster computation. Notably, model optimization takes longer than inference, which typically takes less than a second (refer to Figure \ref{fig:IMG_heatmaps}). 

The \textbf{ROI} analysis for this experiment reveals that all optimization techniques result in a negative ROI, indicating that the costs associated with optimization outweigh the benefits for 100 inferences. 

\begin{table}[ht]
        \caption{Economic impacts of the optimization and inference stages.}
    \label{tab:IMG_cost}
    \begin{tabular}{cccccccc}
         \toprule
         \multicolumn{2}{c}{\multirow{3}{*}{\textbf{Optimization}}} & \multicolumn{3}{c}{\textbf{Total Optimization}} & \multicolumn{2}{c}{\textbf{Inference}} & \multirow{3}{*}{\textbf{ROI}}\\
        \cline{3-7}
         \multicolumn{2}{c}{\multirow{2}{*}{\textbf{Technique}}} & \multicolumn{2}{c}{\textbf{Time (s)}} & \centering \textbf{MOC} & \textbf{MIC} & \centering \textbf{CPI} & \\
         \cline{3-4}
         \multicolumn{2}{c}{} & \textbf{Mean} & \textbf{SD} & \centering \textbf{(€)} & \centering \textbf{(€)} & \centering \textbf{(€)} &\\
        \midrule
         \multicolumn{2}{c}{No Optimization}& 1.20 & 0.58 & 52.36 & 5.07 & 1.24e-04 & -1.000000\\
         \midrule
         \multicolumn{2}{c}{Dynamic Quantization} & 1.94 & 1.07 & 79.72 & 1.23 & 3.0e-05 & -0.999882\\
         \midrule
         \multicolumn{2}{c}{Torch.compile} & 1.48 & 4.55 & 122.33 & 5.35 & 1.32e-04 & -1.000006\\
         \midrule
         \multirow{3}{*}{Local Pruning} & 25\% & 1.26 & 0.47 & 47.50 & 5.10 & 1.26e-04 & -1.000004\\
         &  50\% & 1.35 & 0.83 & 53.67 & 5.10 & 1.26e-04 & -1.000003\\
         &  75\% & 1.37 & 0.90 & 55.15 & 5.08 & 1.25e-04 & -1.000002\\
         \midrule
         \multirow{3}{*}{Global Pruning} & 25\% & 3.93 & 1.95 & 115.68 & 5.08 & 1.25e-04 & -1.000001\\
         &  50\% & 5.46 & 3.06 & 166.22 & 5.08 & 1.25e-04 & -1.000001\\
         &  75\% & 6.98 & 3.99 & 199.14 & 5.09 & 1.25e-04 & -1.000001\\
         \bottomrule
    \end{tabular}
\end{table}

\begin{figure}[h]
    \centering
    \includegraphics[width=0.8\linewidth]{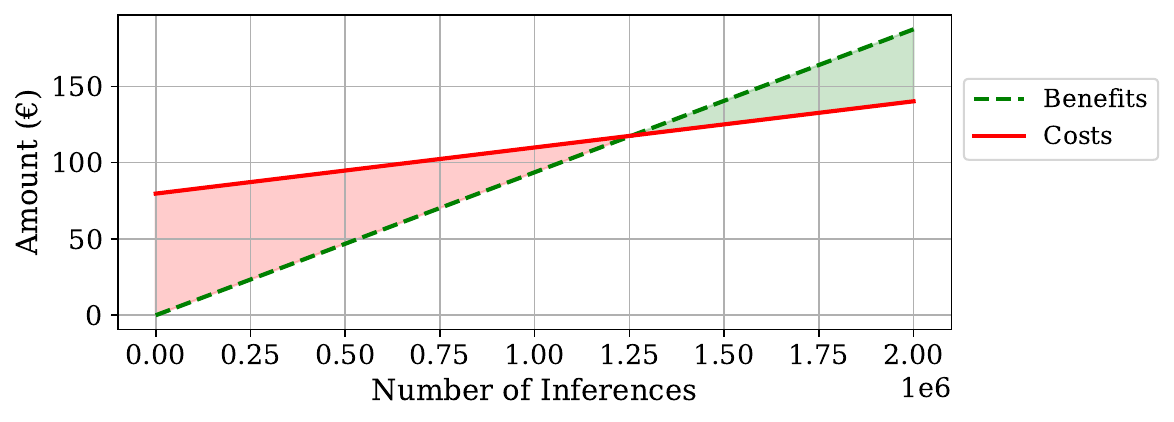}
    \caption{Cost-benefit analysis for dynamic quantization}
    \label{fig:cost_benefits}
\end{figure}

Figure \ref{fig:cost_benefits} illustrates the evolution of costs and benefits associated with dynamic quantization over a range of inferences. While dynamic quantization provides inference savings (as shown in Table \ref{tab:IMG_cost}), it requires 1,256,452 inferences to reach the breakeven point, making it suitable for scalable systems with high inference volumes.

\noindent
\fcolorbox{black}{white}{
    \parbox{0.95\linewidth}{
        \textbf{Finding 2.1}: There is a significant difference between the first inference and the average time taken for the subsequent ones for torch-compiled models.        
    
        \textbf{Finding 2.2}: Despite dynamic quantization being slightly more costly than the baseline in terms of total optimization, it proves cost-effective during the inference phase.

        \textbf{Finding 2.3}: \texttt{Torch.compile} shows significant variation in optimization time and costs, but is not markedly more expensive in inference compared to other methods.

        \textbf{Finding 2.4}: Global pruning methods require longer optimization times, significantly affecting estimated costs.

        \textbf{Finding 2.5}: Dynamic quantization suits scalable systems needing over a million inferences to cover costs.
    }
}

\section{Discussion}
Based on the observations presented in the previous section, this section explores their implications.

\subsection{Implications}
\begin{figure}[!ht]
    \centering
    \includegraphics[width=1\linewidth]{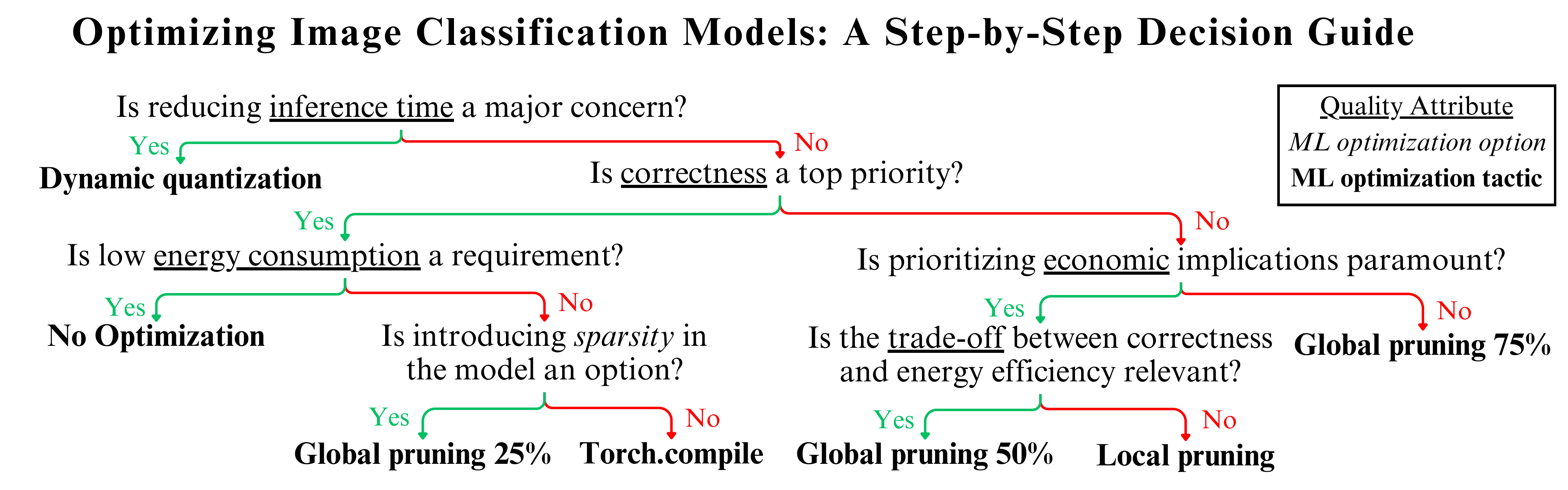}
    \captionsetup{justification=centering}
    \caption{Decision guide for optimizing image classification models based on prioritized quality attributes.}
    \label{fig:IMG_decision_tree}
\end{figure}

\textbf{For ML Engineers}: We present a detailed decision tree in Figure \ref{fig:IMG_decision_tree} to help practitioners understand the effects of each optimization technique and make informed decisions based on prioritized quality attributes. The guide starts by asking whether reducing inference time is a major concern, leading to dynamic quantization, if affirmative. In the case inference time reduction is not prioritized, it further considers the accuracy and the necessity of low energy consumption, recommending no optimization if those are required. Otherwise, global pruning 25\% or \texttt{torch.compile} are recommended, particularly if incorporating sparsity is an option. For scenarios with lower accuracy concerns, recommendations are based on prioritizing economic implications.

\newpage
\textbf{For Hugging Face}:
The examination of Hugging Face models reveals trends in popularity, but many model cards lack crucial details. Training datasets are often undisclosed, and information about model architecture or changes from fine-tuning is frequently missing, limiting reliability.

To address these issues, Hugging Face should enforce detailed documentation standards, including information on training datasets and modifications to fine-tuned models.

\textbf{For PyTorch Libraries}:
For ML optimization, PyTorch libraries would benefit from pruning implementations that not only mask pruned parameters to zero, but actually remove them. This approach would reduce the number of parameters and operations, thereby enhancing model efficiency. Additionally, by eliminating these parameters, the number of computations required would be reduced, leading to faster execution.

\subsection{Related Work}\label{subsec:related_work}

\begin{sidewaystable}
    \caption{Summary of relevant aspects in related work about ML optimization tactics.}
    \label{tab:related_work}
    \begin{tabular}{p{1cm} p{3cm} p{2cm} p{1.7cm} p{2cm} p{1.3cm} p{5.5cm}}
    \hline
    \toprule
    \textbf{Study} & \textbf{ML Optimization} & \textbf{ML Task} \cite{huggingfaceTasksHugging} & \textbf{Datasets} & \textbf{Approach} \cite{githubEmpiricalStandardsdocsstandardsMaster} & \textbf{Libraries} & \textbf{Key Findings} \\
    \midrule
    \cite{tusharma} & Quantization \newline (AWQ, GPTQ, BNB, GGML, GGUF) & Text \newline Generation & wikitext-2 & Benchmarking & Not stated & Energy optimization is key due to varying energy profiles. GGUF, optimized for CPU hardware, shows the importance of co-developing software and hardware.\\
    \midrule
    \cite{DBLP:conf/aaai/LeeJKKP24} & Quantization & Others (Large Language Models) & C4 & Experiment & Pytorch & Development of OWQ, which preserves precision in key weight columns while optimizing the quantization of other dense weights.\\
    \midrule
    \cite{dubhir2021benchmarking} & Quantization (TensorQuant, dynamic \& static quantization) & Graph ML, Image \newline Classification & ImageNet (ILSVRC12), MNIST & Benchmarking & PyTorch, \newline TensorFlow & PyTorch offers detailed modules; TensorFlow supports varied bit representations.\\  
    \midrule
    \cite{gale2019state} & Magnitude pruning, sparse variational dropout, $l_0$ regularization & Translation, \newline Image \newline  Classification &  WMT 2014 English-to-German, ImageNet & Experiment & TensorFlow & Complex techniques yield high compression rates on smaller datasets perform inconsistently, while simple magnitude pruning achieves comparable results. 
    \\
    \midrule
    \cite{DBLP:journals/corr/abs-2404-08831} & Pruning (uniform, non-uniform, iterative) & Image \newline Classification & PanNuke & Experiment & Pytorch & Effective reduction in computation without loss in accuracy.\\
    \midrule
    \cite{9897752} & Pruning, Quantization (PTQ) & Image \newline Classification, Image \newline Segmentation & ISIC & Experiment & EDDL & Introduction of HLSinf, with 90x speedups in medical image applications through pruning and quantization.\\
    \midrule
    \cite{10643325} & Pruning  \newline (Unstructured, \newline Structured, \newline Semi-structured)  & Various \newline (Text, Vision, Multimodal) & Not stated  & Case Survey & Not stated  & Provides a review of pruning methods with a focus on emerging topics like LLMs and vision transformers.  \\ 
    \midrule
    This study & Quantization (dynamic), \texttt{torch.compile}, Pruning (local, global) & Image \newline Classification & ImageNet, Food-101, CIFAR-10 & Experiment & PyTorch & Quantization cuts energy use and time while preserving accuracy, \texttt{torch.compile} balances accuracy and efficiency, and local pruning is the least effective. \\
    \bottomrule
\end{tabular}
\end{sidewaystable}

Table \ref{tab:related_work} summarizes related work. Rajput et al. \cite{tusharma} compared five quantization techniques for text generation inference: Activation-aware Weight Quantization (AWQ), Gradient-based Post-Training Quantization (GPTQ), Bits and Bytes (BNB), GPT-Generated Model Language (GGML), and GPT-Generated Unified Format (GGUF). Additionally, they emphasize the importance of energy optimization. Lee et al. \cite{DBLP:conf/aaai/LeeJKKP24} proposed Outlier-aware Weight Quantization (OWQ) for Large Language Models (LLMs), focusing on preserving precision in weight columns sensitive to quantization errors while optimizing the quantization of other dense weights.  

In the domain of image classification, Rokh et al. \cite{10.1145/3623402} surveyed previous quantization works, showing that QAT achieves higher accuracy than PTSQ due to better compatibility with the quantized model. 

Dubhir et al. \cite{dubhir2021benchmarking} benchmarked PyTorch and TensorFlow quantization for Computer Vision. PyTorch offers more detailed modules and quantization options, but it is limited to int8 for weights. Conversely, TensorQuant allows for more varied bit representations but showed no significant differences in memory usage or testing times. 

Pruning and model compression optimize large pre-trained Deep Neural Networks while compressed convolutional filters and matrix factorization suit training from scratch \cite{goel2020survey}. Adnan et al. \cite{DBLP:journals/corr/abs-2404-08831} show that structured pruning reduces computation in tumor classification, suggesting large models are not always necessary for accurate inference. Flich et al. \cite{9897752} introduced HLSinf for developing neural network accelerators on FPGAs using EDDL (European Distributed Deep Learning Library). It enhances inference efficiency in through pruning and quantization, achieving up to 90x speedups. 

Reguero et al. \cite{REGUERO2025103906} addressed environmental concerns in DL by introducing a prediction model that reduces energy consumption and improves accuracy through strategic application of layer freezing, quantization, and early stopping during training.

Our study builds on previous research by examining how quantization, pruning, and \texttt{torch.compile} affect not only model performance and resource usage, but also the economic costs associated with optimization and inference. While earlier studies have focused on accuracy or computational efficiency, we highlight the economic implications.

\section{Threats to Validity}
In this section, we address potential threats to the study's validity, aiming to clarify the constraints and biases that could affect the interpretation of the findings. 

\textbf{Internal Validity}: 
There is a dependency on the chosen models due to their reliance on the models available on Hugging Face, some of which had to be excluded due to missing files or incomplete training information. As a mitigation strategy, we have implemented a stratified sampling based on model popularity and size to ensure a representative selection. Although certain optimization techniques, like static quantization, were initially considered but found unsuitable for the diverse range of models and architectures studied. Furthermore, the study assumes accurate data from the wattmeter, nvidia-smi, and EnergiBridge, but inconsistencies or failures could impact measurement quality.

\textbf{External Validity}: Differences in hardware specifications may lead to varying metrics, potentially affecting study reproducibility, though not expected to impact conclusions. Additionally, there exist pre-trained ML models with different architectures from those used in this study, which may result in variations in performance. Updating the pipeline with new models can serve as a mitigation strategy.

\textbf{Construct Validity}: A potential threat could arise from data overlap if experiments were run concurrently. To mitigate this risk, experiments were conducted sequentially, minimizing the influence of confounding variables. This approach ensured that each measurement of key quality attributes was precise and distinct. Rigorous control over experimental conditions was maintained throughout.

\textbf{Conclusion Validity}: The absence of standardized reporting practices concerning model cards may have influenced the conclusions drawn from the analysis of datasets and models on Hugging Face. Moreover, the current implementation of pruning in PyTorch sets weights to zero rather than removing them entirely. As a result, this method does not significantly reduce the number of parameters or the computational operations required, which is typically the primary objective of pruning.

\section{Conclusions and Future Work}
This study explores model optimization techniques for pre-trained ML models. The analysis of image classification datasets available on Hugging Face underscores the prevalence of classic Computer Vision datasets and their versatility across multiple ML tasks. 

Among optimization strategies, dynamic quantization emerges as a standout, showcasing significant time and energy consumption reductions while preserving accuracy. Additionally, \texttt{torch.compile} demonstrates a balance between accuracy and energy efficiency. However, local pruning fails to yield positive effects on performance. 

By shedding light on the effectiveness of diverse optimization methods, this study provides practitioners with crucial insights for informed decision-making throughout the model development and deployment processes.

Future work spreads in several directions. While this study provides valuable insights into image classification model optimization, expanding its scope to include NLP models' inference efficiency and multimodal applications (integrating text, image, and audio data) would provide a comprehensive view across different domains and real-world scenarios. Additionally, exploring other optimization libraries, such as TensorFlow, could further offer new perspectives on enhancing model efficiency.

The extensive database created during this study can serve as a foundation for categorizing models, by clustering them according to energy consumption and performance metrics. These insights can be used to craft customized energy labels for ML model inference, which could be integrated into GAISSALabel tool \cite{duran2024gaissalabel}.

\section*{Acknowledgment}
This work has been partially funded by the Spanish research projects GAISSA (TED2021-130923B-I00 by
MCIN/ AEI/10.13039/501100011033) and DOGO4ML (ref. PID202 0-117191RB-I00).

\bibliography{references.bib}

\end{document}